\documentclass[10pt,twocolumn,letterpaper]{article}
\pdfoutput=1
\usepackage{iccv}
\usepackage{times}
\usepackage{epsfig}
\usepackage{graphicx}
\usepackage{amsmath}
\usepackage{amssymb}
\usepackage{booktabs}
\usepackage{multirow}
\usepackage{epstopdf}

\usepackage[pagebackref=true,breaklinks=true,letterpaper=true,colorlinks,bookmarks=false]{hyperref}

\iccvfinalcopy 

\def\iccvPaperID{9589} 

\ificcvfinal\fi

\begin{document}
	
\title{Countering Adversarial Examples: Combining Input Transformation and Noisy Training}

	\author{Cheng Zhang\\
		Nanjing University of Aeronautics and Astronautics\\
		Nanjing\\
		{\tt\small zhang927566@nuaa.edu.cn}
		\and
		Pan Gao\\
		Nanjing University of Aeronautics and Astronautics\\
		Nanjing\\
		{\tt\small Pan.Gao@nuaa.edu.cn}
	}
	
\maketitle	

\begin{abstract}
	
	Recent studies have shown that neural network (NN) based image classifiers are highly vulnerable to adversarial examples, which poses a threat to security-sensitive image recognition task. Prior work has shown that JPEG compression can combat the drop in classification accuracy on adversarial examples to some extent. But, as the compression ratio increases, traditional JPEG compression is insufficient to defend those  attacks but can cause an abrupt accuracy decline to the benign images.  In this paper, with the aim of fully filtering the adversarial perturbations, we firstly make modifications to traditional JPEG compression algorithm which becomes more favorable for NN. Specifically, based on an analysis of the frequency coefficient, we design a NN-favored quantization table for compression. Considering compression as a data augmentation strategy, we then combine our model-agnostic preprocess with noisy training. We fine-tune the pre-trained model by training with images encoded at different compression levels, thus generating multiple classifiers. Finally, since lower (higher) compression ratio can remove both perturbations and original features slightly (aggressively), we use these trained multiple models for model ensemble. The majority vote of the ensemble of models is adopted as final predictions. Experiments results show our method can improve defense efficiency while maintaining original accuracy.
	
\end{abstract}

\section{Introduction}

\label{sec:intro}
Adversarial attack presents a major challenge for the prevalent deep neural networks used for image classification and recognition \cite{AE}. Several countermeasures have been proposed against adversarial examples, mainly including 
model-specific hardening strategies and model-agnostic defenses. Typical model-specific solutions like ``adversarial training" \cite{AEtraining,dd,sat,at_fast,at_free} can rectify the model parameters to mitigate the attacks by using the iterative retraining procedure or modifying the inner architecture. However, it is generally believed that, network's architectural elements would matter little unless making them larger and deeper in improving adversarial robustness. In contrast,  model-agnostic solutions like input dimension reduction or direct JPEG compression \cite{jpg1,jpg2}, become more feasible and practical, which attempt to remove adversarial perturbations by input transformations before feeding them into neural network classifiers.

\begin{figure}
	\centering
	\includegraphics[scale=0.55]{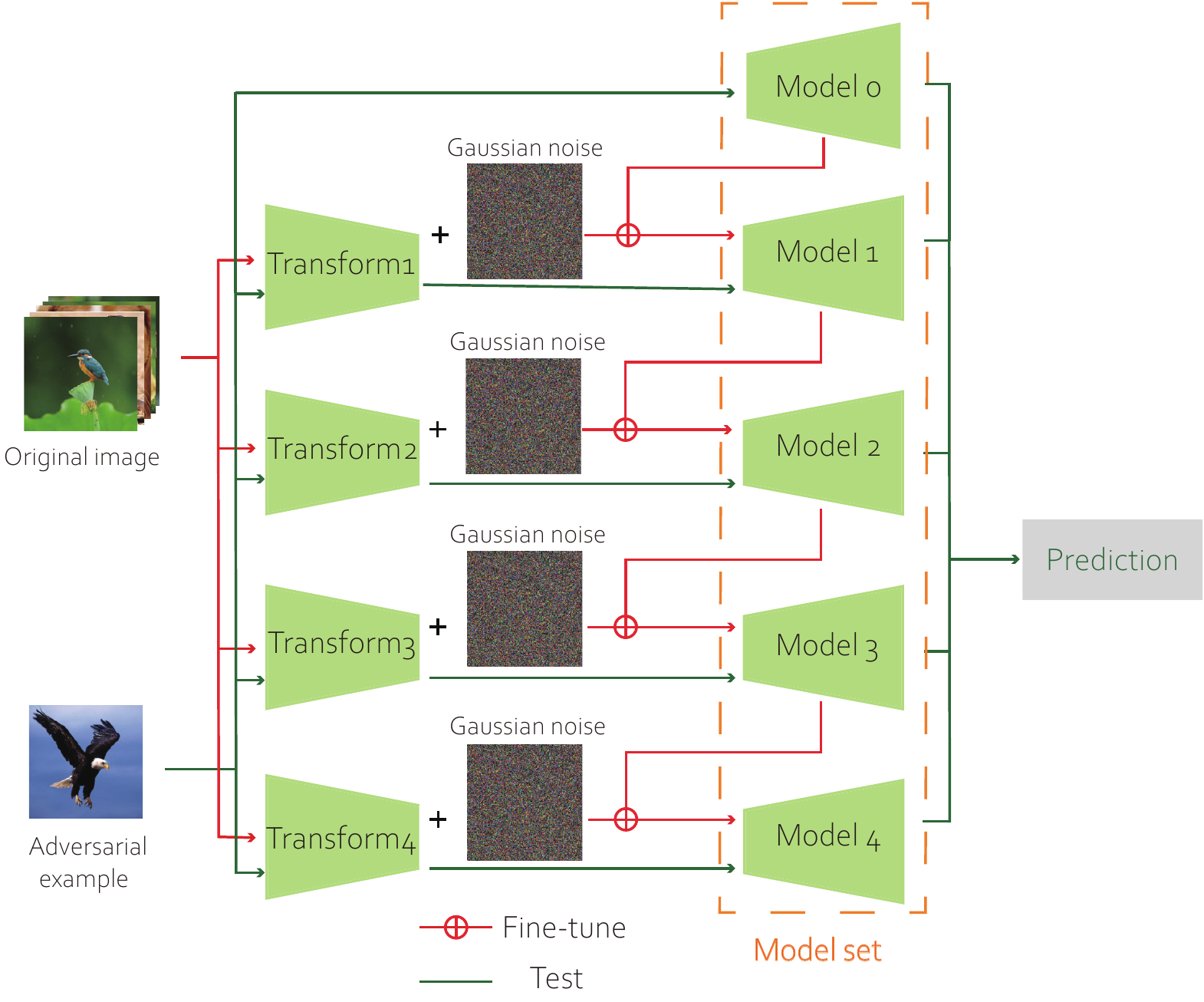}
	\caption{Overview of our combination method. Different transform modules represent different level of compression for the original images. The initial model, i.e., model 0, is the pre-trained model on benign images of ImageNet dataset, such as ResNet or Inception-v4. }
	\label{fig:00}
	\vspace{-3mm}
\end{figure}


For mitigating adversarial examples, standard JPEG compression has been explored in \cite{jpg1,jpg2}.  But, in these works, they have shown that JPEG cannot achieve a good balance between countering adversarial examples and classifying benign images, \emph{i.e.}, lower quality factor (QF) for JPEG compression achieves better defense efficiency but causes a significant feature loss on benign images. To resolve this problem, we first optimize the JPEG based transformation process in this work, to improve defense efficiency against adversarial examples and maintain classification accuracy on benign images. 
Firstly, we analyze the distributions of the DCT coefficients for 6 color channels (i.e., R, G, B, Y, Cb, Cr) on both benign images and polluted images to find out adversarial perturbations' distribution at all 64 frequency bands. With the frequency analysis, we then divide the frequency coefficients into two types, i.e., the original favored (OF) band and the adversarial favored (AF) band. Finally, the corresponding defensive quantization parameters for these two bands are derived, where the number of the DCT coefficients that should be included in each type of band is jointly optimized.

With the purpose to further  achieve both accuracy and robustness, we fine-tune the model with our own pre-processed images. Firstly, as a data augmentation, training images will be compressed using our compression algorithm. This idea shares the similar spirit of the  image cropping-rescaling method proposed in \cite{tvm&quilting}, in which, the neural network re-trained on randomly cropped-rescaled images yields better performance than other input transformations. 
Secondly, Gaussian noise is added to the compressed images to mimic the adversarial perturbations (detailed in Section \ref{ae_training}), which is based on the fact  that strong adversaries are not necessarily needed during adversarial training as demonstrated in  \cite{at_fast}. 
However, as the model is commonly noisy trained with compressed images of a certain level of quality, there still exists unavoidable trade-off between robustness and accuracy. To achieve the best balance, we generate a number of classifiers by fine-tuning the neural network using a variety of degrees of compression quality images during aforementioned pre-processing. After having obtained a set of classifiers, the final prediction value if chosen to be the label  maximizes the average confidence (i.e., the output of Softmax layer) of each classifier.

Figure \ref{fig:00} shows an overview of our overall method, where we combine the input transformation and the noisy training. The pre-processing is implemented by using a proposed compression followed by adding the general  Gaussian noise. The model set is realized by fine-tuning the models with compressed images at different compression level. The initial model used for fine-tuning is the pre-trained model on benign images. The models retrained with different compression levels are ultimately utilized together in an ensemble defense. Experimental results demonstrate the defense efficiency and the legitimate classification efficiency of the proposed algorithm against a variety of adversarial examples in the gray-box, black-box and white-box scenarios. 
The implementation code of the algorithm proposed   will be made publicly available.

\section{Related Works}
\label{sec:Related work}

\subsection{Adversarial Attacks}
One of the first and simple but quite effective attack is the \textbf{Fast gradient sign method} (FGSM \cite{fgsm}). It simply takes the sign of the gradient of loss function \textit{J} (e.g., cross-entropy loss) w.r.t the input image $x$ and multiply with magnitude $\epsilon$ as perturbations,
\begin{equation}
x_{adv} = x + \epsilon \cdot sign( \bigtriangledown_{x}J(\theta,x,y)).
\end{equation}
where $\theta$ is the parameters set of neural network and $y$ is the ground truth label of $x$. The parameter $\epsilon$ is the magnitude of perturbation which controls the similarity of adversarial examples and original image. By trying to find a high success rate adversarial example but has as small dissimilarity with original image as possible, \cite{bim} proposed an iterative version of FGSM, \textbf{I-FGSM}. It iteratively applies FGSM in every iteration and clips the value to ensure per-pixel perturbation below attack magnitude, 
\begin{equation}
x^{(i)}_{adv} = Clip_{x,\epsilon}\{x^{(i-1)}_{adv} + \epsilon \cdot sign( \bigtriangledown_{x}J(\theta,x,y))\}.
\end{equation}
\textbf{Deepfool} \cite{deepfool} assumes neural network as a linear classifier, and then projects $x$ onto a linearization of the decision boundary. The distance from $x$ to decision boundary is computed as perturbation. Since the assumption aforementioned is  overly simplistic, Deepfool keeps iterating until it finds a success adversarial example.
\textbf{Carlini-Wagner' $L_{2}$ attack}(\textbf{C\&W$L_{2}$})  \cite{cw} is an optimization-based attack that adds a relaxation term to the perturbation minimization problem based on a differentiable surrogate of the model. The optimization problem are minimizing
\begin{equation}
\parallel x - x'\parallel+ \lambda_{max}(-\kappa, Z(x')_{k}- max(Z(x')_{k'}:k'\neq k)) 
\end{equation}
where $\kappa$ controls the confidence of predictions made by neural network, and $Z(\cdot)_{k}$ represents the logits value (input for softmax layer) corresponding to class \textit{k}.
\textbf{BPDA} \cite{bpda} recurrently computes the adversarial gradient after applying defense:
\begin{equation}
x_{adv} = Clip(x + \epsilon \cdot sign(\bigtriangledown J_{\theta , Y}(Def(x))))
\end{equation} 
where $J$ represents the loss function of classifier and $Def()$ is the used defense method.
\subsection{Model-Agnostic Defenses}

\begin{figure*}[t]
	\centering
	\includegraphics[scale=0.4]{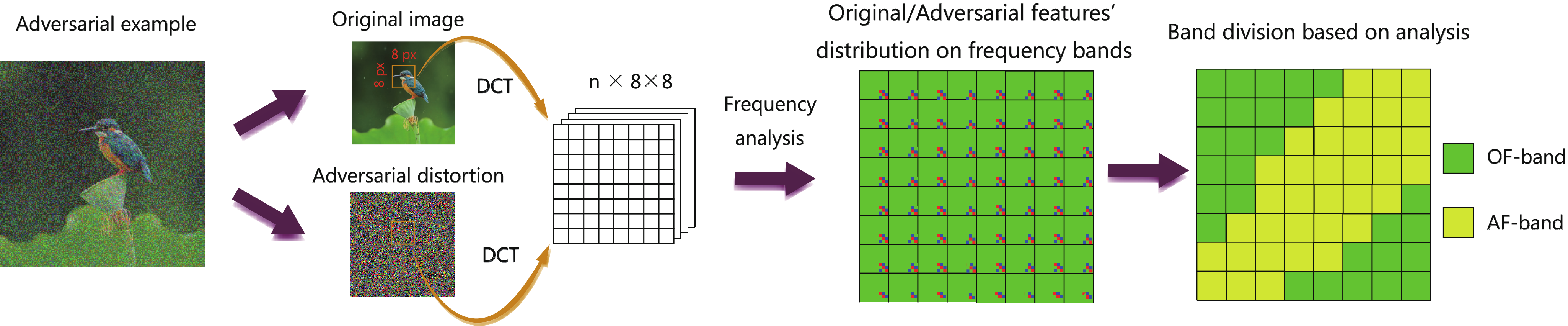}
	\caption{Analysis of original and adversarial features' distribution at frequency domain and band division.}
	\label{fig:02}
	\vspace{-2mm}
\end{figure*}

Recently, \cite{deepn} developed a deep neural network favorable JPEG-based image compression framework called Feature distillation which preprocesses images using modified quantization table in JPEG. \cite{tvm&quilting} combined input transformations like TVM \cite{tvm}, image quilting \cite{quilting} and image cropping. However, both TVM and image quilting are time-consuming. \cite{jpg1} empirically reported that JPEG compression can reverse only small perturbations, but the reason behind is not explained.  \cite{jpg2} proposed a JPEG compression based ensemble method, called ``vaccinating", to mitigate adversarial attacks by voting the results based on a variety of compression rates.
\cite{fd} proposed an adversary-concerned  JPEG compression framework, which modified quantization table to eliminate the malicious perturbations. However, it analyzed the  DCT coefficients of all channels as a whole, and used a heuristic design flow for quantization. In this work, we consider an analysis of the frequency of both the original images and perturbations on different channels, and design a more detailed optimization solution. 
\subsection{JPEG Compression}
JPEG \cite{jpeg}, \cite{li2020optimizing} is one of the most popular lossy compression standards for digital images. First, the image is converted from RGB into a different color space called YCbCr. 
After color space transformation, each channel is split into $8\times8$ blocks. Each  block of each component  is converted to a frequency-domain representation, using discrete cosine transform (DCT) to obtain 64 coefficients.  The coefficients are quantized by  a  quantization table \cite{jpeg}. The table is designed to preserve the low-frequency components and discard high-frequency details, because human visual system (HVS) is less sensitive to the information loss in high-frequency  bands. After quantization, all the quantized coefficients are ordered into the zig-zag sequence. The differential coded DC and AC coefficients will be further compressed by entropy coding. A reversed procedure of aforementioned steps can decompress an image.

\section{Our Proposed Approach}

In the following,  we firstly analyze the DCT coefficient distribution on the respective three channels of the two color formats, i.e., RGB and YCbCr. Then, we derive the best quantization steps for defending perturbation. Finally, we propose a noisy-based adversarial training based robustness reinforcement method. To further enhance defense efficiency, we propose an ensemble method.

\subsection{Frequency Analysis}
Standard JPEG compression is based on two assumptions, i.e.,  HVS are more sensitive to low frequency components of DCT coefficients and brightness channel than high frequency component and  chrominance channels, respectively. However,  neural networks learn features in a quite different way. As shown in the prior work \cite{dctcoefficient}, the image feature is highly related to the standard deviations ($\delta_{i,j}$) of the coefficient, and  a larger $ \delta_{i,j} $ means more features in band $(i,j)$ can be learned by NN classifiers. Inspired by this, we analyze the distribution for each frequency component of the channels of the RGB and YCbCr.   

Our frequency analysis is illustrated in Figure \ref{fig:02}. We select 10k original images from ImageNet randomly and then use them to generate 10K adversarial examples. Each of adversarial examples can be split into original image and adversarial distortion which would be represented as both the RGB and YCbCr spaces. Therefore, each image (original image or adversarial distortion)  can be split into 6 color channels (i.e., R, G, B, Y, Cb, Cr). Each separate channel is then partitioned into $8\times8$ blocks, followed by a block-wise DCT. The standard deviations of each frequency channel are computed. This statistical information can tell us the distribution  of the original/adversarial features on frequency bands. Based on that distribution, we can optimize quantization table.

\par
\begin{figure}[t]
	\centering
	\includegraphics[scale=0.5]{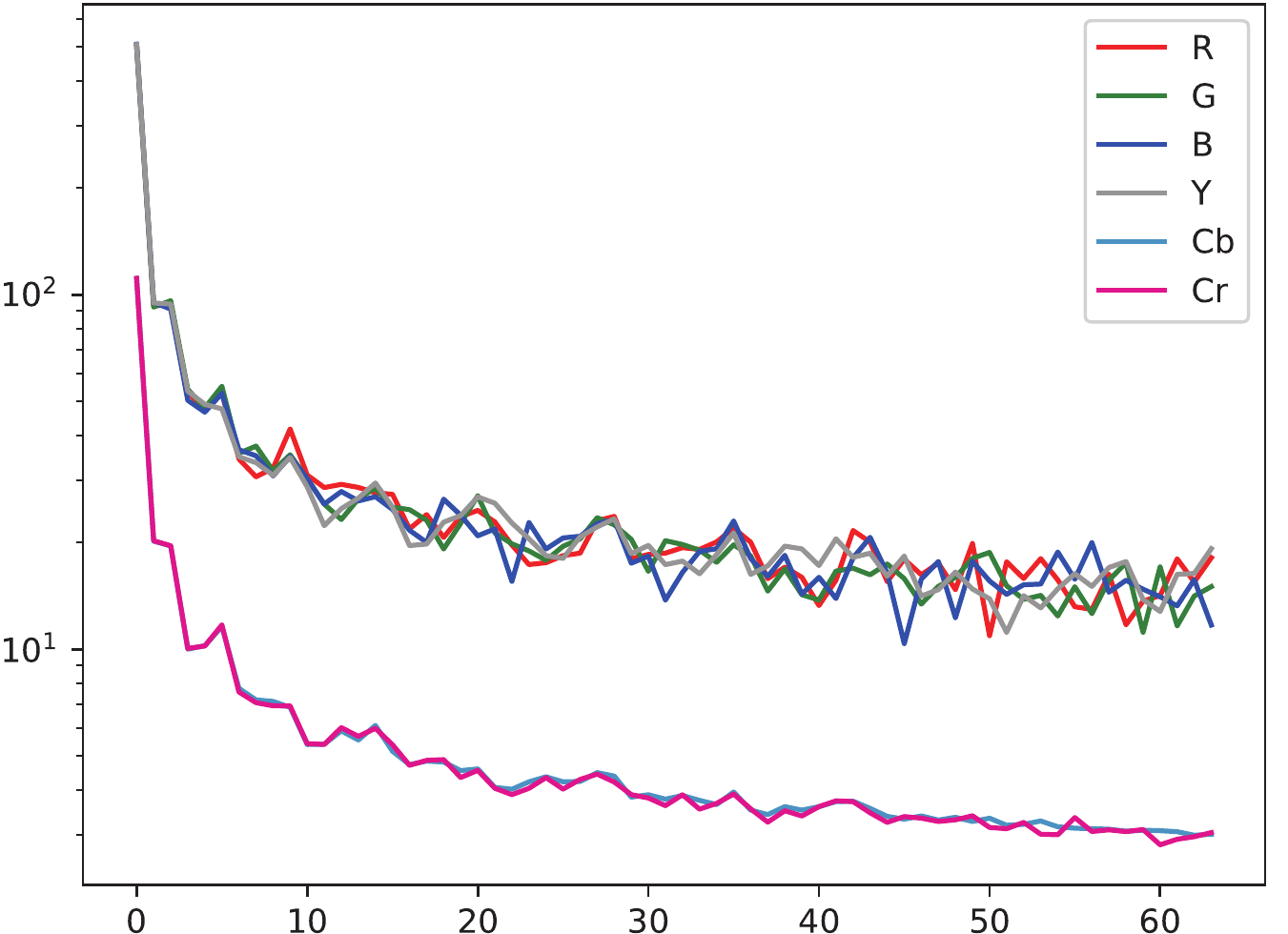}
	\caption{Statistical information about frequency of 64 component in different color channels. The \emph{y}-axis represents the standard deviation ($\delta_{i,j}$) and the \emph{x}-axis represents the 64 components.}
	\label{fig:03}
\end{figure}
\begin{figure}[t]
	\centering
	\includegraphics[scale=0.5]{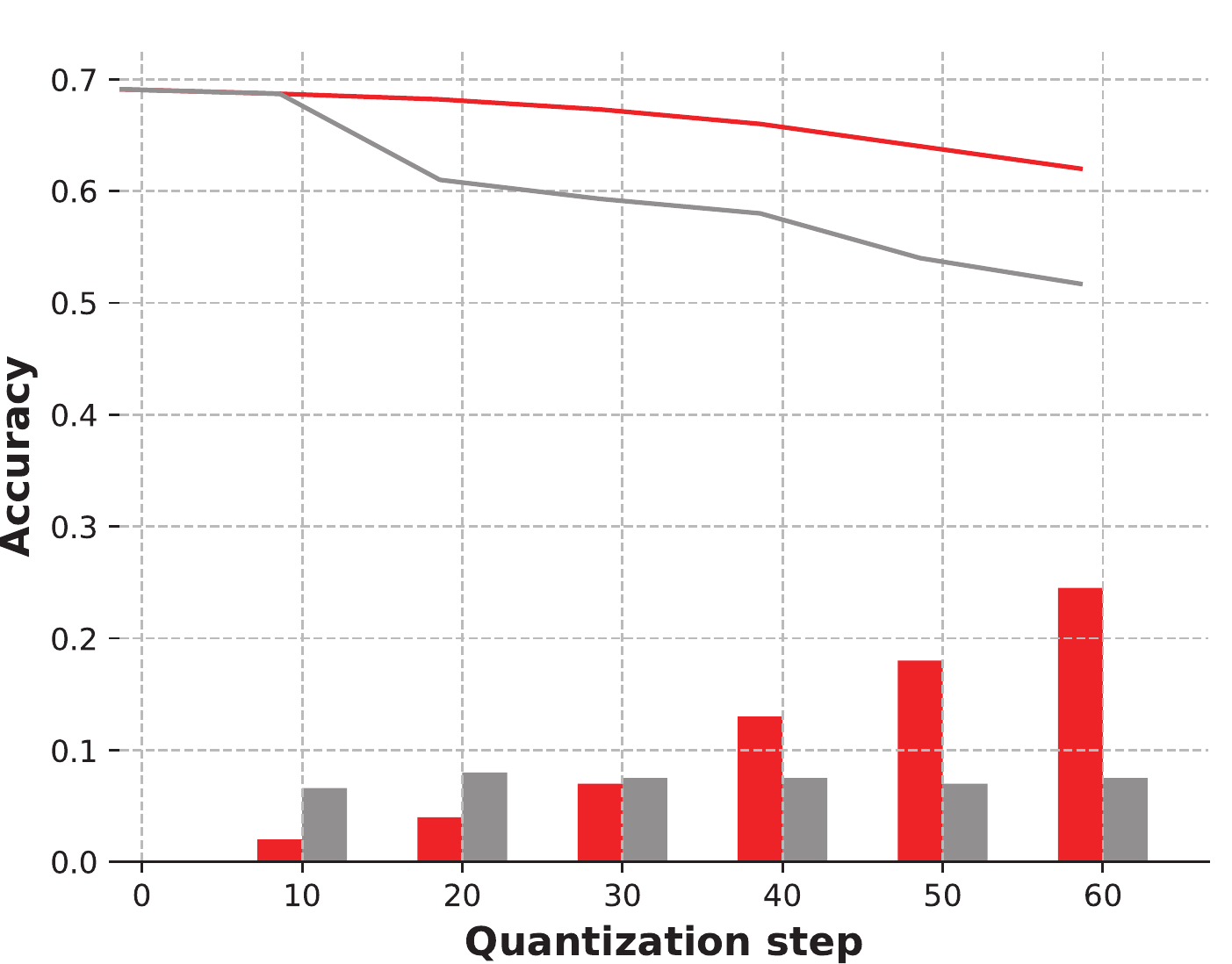}
	\caption{Comparison between doing quantization process at the RGB  and YCbCr domains. The lines represent the accuracy on preprocessed original image (red for applying quantization at RGB domain, gray for YCbCr), and the histograms represent the accuracy on preprocessed adversarial examples (red for applying quantization at RGB domain, gray for YCbCr). Here, adversarial examples are generated using FGSM ($\epsilon$ = 0.008) and tested on ResNet-50.}
	\label{fig:04}
\end{figure}

As our experimental results in Figure \ref{fig:03} show, RGB space has almost the same $\delta_{i,j}$ for the frequency components among the three channels. However, Cb, Cr channel in YCbCr space has a significantly less $\delta_{i,j}$ in comparison to each RGB channel, though Y channel aligns with RGB channel in terms of $\delta_{i,j}$. That is caused by the down-sampling step in standard JPEG process.
According to previous conclusion that higher $\delta_{i,j}$ means more features, we consider the  RGB to YCbCr color space transformation, would induce feature loss in the following quantization step so that there would be an accuracy decline when classified by neural network. To validate the hypothesis above, we use the same quantization table for compression in RGB and YCbCr domains. Results are shown in Figure \ref{fig:04}. Though the defense efficiency is slightly worse when QS (quantization step) is less than 20, as QS increases, directly doing DCT in RGB domain not only can preserve a higher accuracy on benign images but also have nearly 2 times larger defense efficiency than YCbCr domain. 
In addition, another advantage of employing RGB space  is that, since RGB space has the same statistical characteristics for the respective three channels, it allows us to design only one quantization table for the lossy coding of the DCT coefficients. 

\subsection{Quantization Table Design}
\label{sec: table design}
In this section, we redesign the quantization table for the purpose of removing the perturbation in the image.

\textbf{Step 1.}
As also shown in \cite{dctcoefficient}, the unquantized coefficients can be approximated as normal distribution with zero mean but different standard deviations ($\delta_{i,j}$). Since the adversarial example is a linear combination of two normal-distributed variables, i.e.,  original DCT coefficient and the perturbation, its $\delta$ can be also considered as a linear combination of the $\delta s$ of these two variables. 
Since $\delta_{i,j}$ of adversarial example is a linear summation of original image and perturbations, this statistical information can tell us the relationship between each frequency component and features. 
Assume that $\delta_{i,j}^{b}$ indicates $\delta_{i,j}$ of DCT coefficients for the benign image, and $\delta_{i,j}^{a}$ indicates $\delta_{i,j}$ of DCT coefficients for adversarial distortions.
We then import $\delta_{i,j}^{'}$=$\delta_{i,j}^{a}$/$\delta_{i,j}^{b}$  to characterize correspondence between two contrary features for the frequency component.
The  $\delta_{i,j}^{'}$ is thus a measure of how many perturbation is introduced into the current considered frequency component, i.e., coefficients of those frequency components with larger $\delta_{i,j}^{'}$ contains greater proportion of adversarial features. With the $\delta_{i,j}^{'}$ calculated for all the coefficients,  we sort the DCT coefficients in a way that the associated  $\delta_{i,j}^{'}$ increases. After the ascending sorting for the DCT coefficients,
we can partition 64
coefficients to OF (origial-favored) band and AF (Adversarial-favored) band. The number of DCT coefficients that is included in each type of band is determined by the optimization algorithm, which will be detailed in the following. 

\begin{figure}[t]
	\includegraphics[scale=0.2]{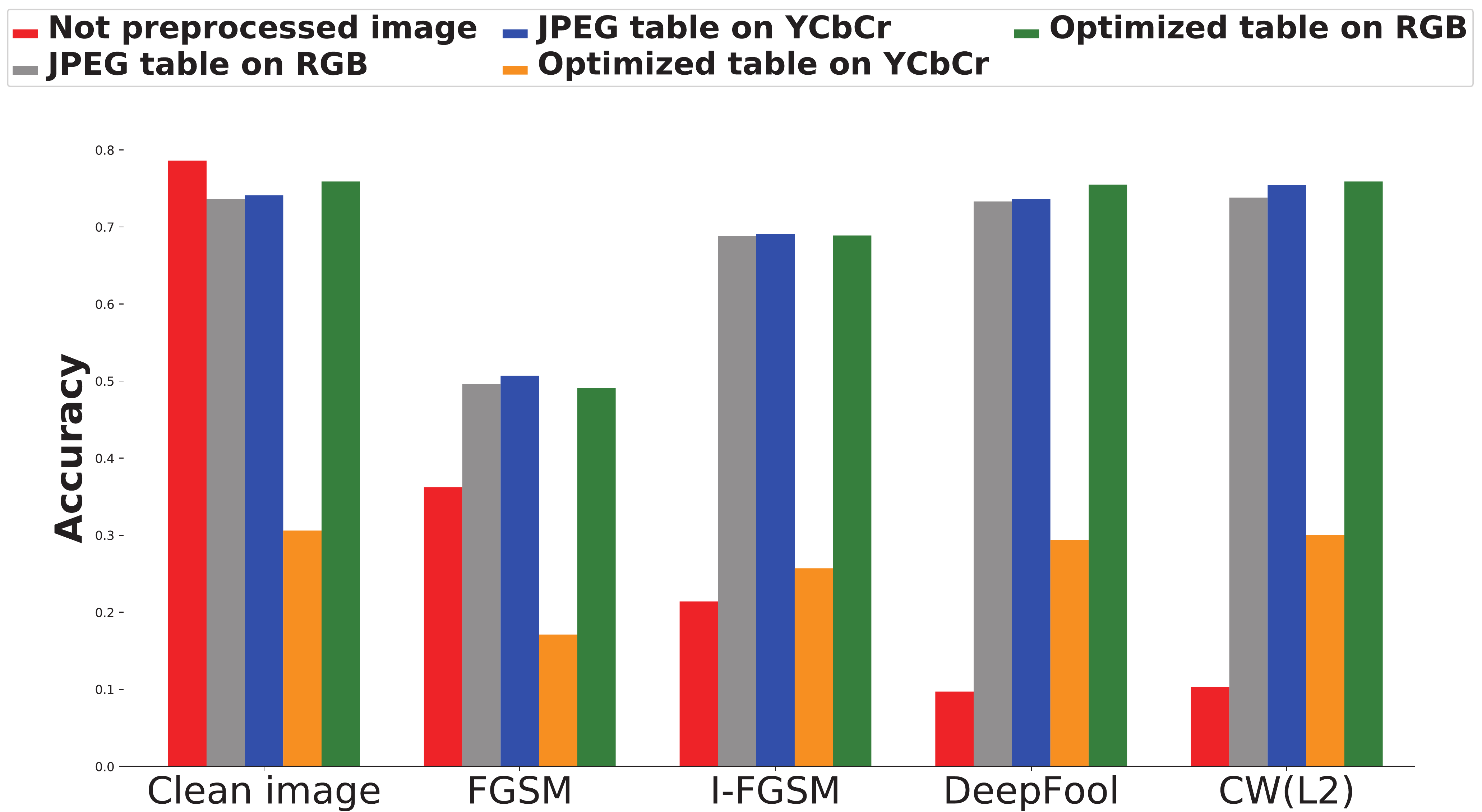}
	\caption{Ablation study on how the quantization table and color transform contribute to robustness. Clean images are from ImageNet validation set.}
	\vspace{-2mm}
	\label{fig:05}
\end{figure}

\textbf{Step 2.}
For these two types of bands, we use different quantization parameters to balance the defense efficiency and the testing accuracy of the legitimate examples. 
When adversarial distortion $\rho_{x}$ is added  with intensity $\epsilon$ ($\epsilon=\rho_{x}/255$) into original $8\times8$ block ${x}$, the formed adversarial block can be represented as:
\begin{equation}
x_{adv}=x+\rho_{x}=(x/255+\epsilon)\times 255
\end{equation}

Input block will be linearly separated by 2D-DCT transformation as:
\begin{small}
	\begin{equation}
	\label{eqn:DCT}
	DCT(x_{adv})=DCT(x)+DCT(\rho_{x})=C_{x}\cdot B+C_{\rho_{x}}\cdot B
	\end{equation}
\end{small}
In  \eqref{eqn:DCT}, $C_{x}$ and $C_{\rho_{x}}$ are DCT coefficients, and $B$ denotes the DCT basis function. 
The maximum magnitude of $C_{\rho_{x}}$can be calculated by the sum of all 64 frequency components and each term is bounded by \textit{$cos(\theta)\cdot\epsilon$}, where $\theta$ is dependent on the DCT basis generating function \cite{jpeg}. Thus we have $-8 \cdot\epsilon<C_{\rho_{x}}<8\cdot\epsilon$. Generally speaking, the  quantization and dequantization steps in JPEG provides an opportunity for filtering adversarial perturbations. If we want to eliminate the perturbation $C_{\rho _{x}}$, we need a QS to satisfy:
\begin{small}
	\begin{equation}
	Round((C_{x}+C_{\rho_{x}})/QS)\times QS\approx Round((C_{x})/QS)\times QS
	\end{equation}
\end{small}
Noting such a process cannot fully recover the original value, even though the perturbation is removed. Since $(Round(C_{x})/QS) \cdot QS\ne C_{x}$, let $\eta=Round(C_{x})/QS$, we can get $\hat\theta=|C_{x}-\eta \cdot QS|$, which is the remainder of $C_{x}/QS$. With these notations, we have the following:

$Round(C_{x}/QS)=(C_{x}-\hat\theta)/QS$\\

\begin{small}
	$Round((C_{x}+C_{\rho_{x}})/QS)=Round((C_{x}-\hat\theta+\hat\theta+C_{\rho_{x}})/QS)$\\
\end{small}
\begin{equation}\label{eqn:zero}
Round((C_{x}+C_{\rho_{x}})/QS)=\eta+Round((\hat\theta+C_{\rho_{x}})/QS)
\end{equation}\label{perturbation_removal}
As can be observed from the above derivation,   either too small or too large QS can induce large rounding error in the quantization and dequantization processes. Thus, in order to make the second term in the right-hand side of \eqref{eqn:zero} equals zero, we choose QS=16$\cdot\epsilon$, which is used for quantizing the OF bands to preserve original features.

\begin{figure*}[t]
	\centering
	\includegraphics[scale=0.45]{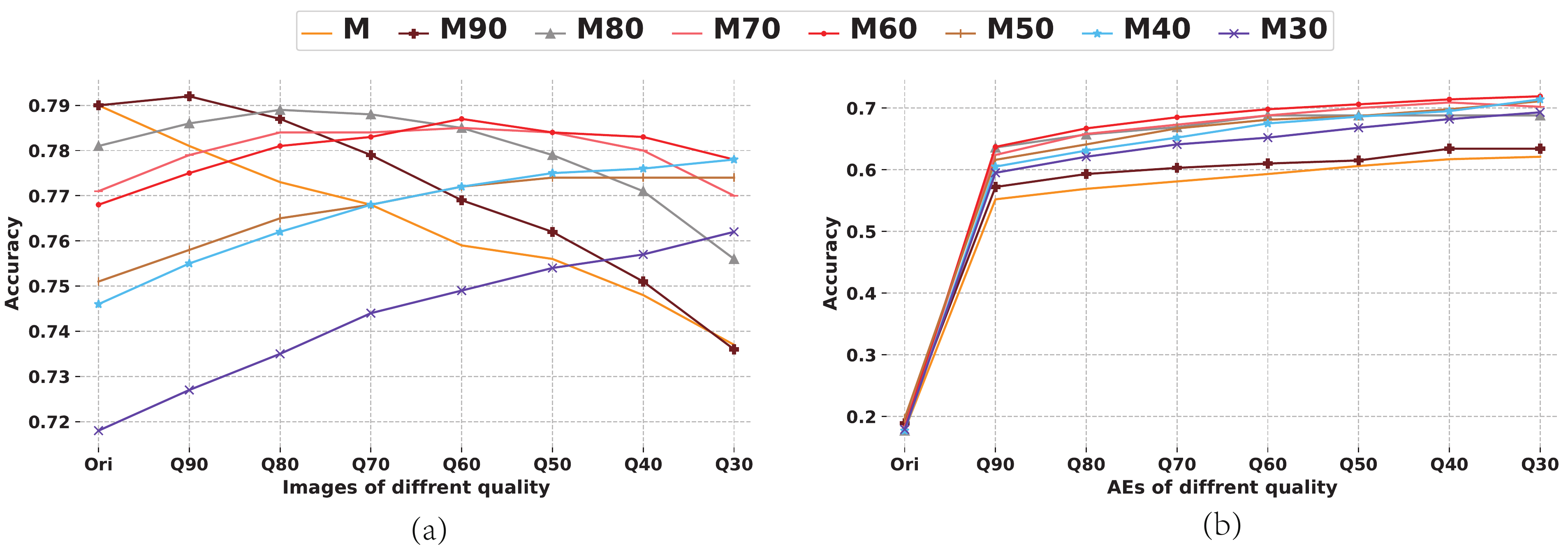}
	\caption{The performance of each individual model, each model is fine-tuned using certain but different quality images: (a) Benign images from ImageNet validation set are compressed at different quality and then tested on each model. (b) Adversarial examples (AEs) generated using I-FGSM ($\epsilon=0.008$) are compressed at different quality and then tested on each model.
		Here \emph{x}-axis represents images at 8 different compression qualities, and \emph{y}-axis represents the test accuracy.	
	}
	\label{fig:06}
	\vspace{-3mm}
\end{figure*}

\par
\textbf{Step 3.} For determination of the number of OF bands in the partition process and the  QS for AF bands, we develop an iterative algorithm. In general, more DCT coefficients included in the OF bands can lead to better classification accuracy on the benign images. But,  it will result in degraded defense efficiency for the polluted images. To better balance the classification accuracy and the defense efficiency,  we partition the  DCT coefficient-sorted block according to the right diagonal line following the zigzag scanning order  starting from the DC coefficient. The top-left corner DCT coefficients belong to  OF bands, while the remaining are AF bands. As there are 15 diagonal lines for a $8\times8$ block, we have a total of 15 different partition patterns (P\_Pattern). To determine which P\_Pattern is appropriate for against adversarial attacks, we test the defense efficiency and the classification efficiency for each P\_Patterns. During each test, we also conduct an exhaustive search of the QS for AF (QS\_AF) bands given the QS for OF (QS\_OF) being 16$\times255\epsilon$. The search range of the QS\_AF is from 1 to 121 with step size of 5. We repeat the P\_Pattern search and the QS\_AF search until the defense efficiency converges and the decrease of the classification efficiency on the benign image is no more than 1\%. The detailed implementation of the proposed iterative algorithm  is summarized in Algorithm 1. Based on our quantization table optimization algorithm, we adopt 16 as QS\_OF for OF frequency bands, 50 as QS\_AF for AF frequency bands.

\begin{table}[t]
	\begin{tabular}{l}
		\toprule
		\textbf{Algorithm 1}: Iterative algorithm to optimize QS\\
		\midrule
		1\hspace{0.2cm}QS$\_$AF = 1\textcolor{blue}{$\#$Initial QS for AF bands}\\
		2\hspace{0.2cm}QS$\_$OF = 16\textcolor{blue}{$\#$QS for OF bands, we adopt $\epsilon$=0.004} \\
		3\hspace{0.2cm}for \emph{k} in range(1,15):\\
		4\hspace{0.4cm}Initialize P\_Pattern[\emph{k}] \\
		5\hspace{0.2cm}\textcolor{blue}{$\#$select 1000 benign images}\\
		6\hspace{0.2cm}X$\_$ben = Dataset(ImageNet,1000)\\
		7\hspace{0.2cm}\textcolor{blue}{$\#$generate 200 AEs from correctly classified images}\\
		8\hspace{0.2cm}X$\_$adv = Adv(X$\_$benign,200)\\
		9\hspace{0.2cm} for QS$\_$AF in range(1,121,5) and \emph{k} in range(1,15): \\
		10\hspace{0.7cm}        X$\_$ben$\_$de = update$\_$jpeg(X$\_$ben, \emph{k}, QS$\_$AF, QS$\_$OF)\\
		11\hspace{0.7cm}        X$\_$adv$\_$de = update$\_$jpeg(X$\_$adv, \emph{k}, QS$\_$AF, QS$\_$OF)\\
		12\hspace{0.7cm}        model.predict(X$\_$ben$\_$de,~X$\_$adv$\_$de)\\
		13\hspace{0.2cm}\textcolor{blue}{$\#$Accuracy decline on benign images}\\
		14\hspace{0.7cm}        Acc$\_$dec = Cal$\_$acc$\_$dec()\\
		15\hspace{0.2cm}\textcolor{blue}{$\#$Defense efficiency on adversarial examples}            \\ 
		16\hspace{0.7cm}        Def = Cal$\_$def()\\
		17\hspace{0.2cm}      [QS$\_$AF*,~\emph{k*}] = ${\rm{argmax}}_{[QS\_AF,~k]}(Def)$\\    \hspace{4.5cm} ~~s.t. ~~$Acc$\_$dec$$<$1$\%$              \\   
		18\hspace{0.2cm}Return P\_Pattern[\emph{k*}],  QS$\_$AF*        \\        
		
		\bottomrule
	\end{tabular}
\vspace{-5mm}
	\label{tab:00}
\end{table}

\subsection{Ablation Study}
As the optimized quantization table is designed, we conduct the ablation survey to figure out how skipping color transform and optimized quantization table contributes to the robustness. We consider 4 combinations: (1) traditional JPEG (original quantization table + color transform), (2) original quantization table + skipping color transform, (3) optimized quantization table + color transform, (4) optimized quantization table + skipping color transform. Four attacks aforementioned are adopted for testing on Inception-v4 \cite{incep}. As Figure \ref{fig:05} shows, if we maintain the quantization table of traditional JPEG, skipping the color transform process can outperform on both benign image and adversarial examples. When applying our optimized quantization table, accuracy drop drastically with color transform. However, our optimized quantization table utilized on RGB space has better defense efficiency than the first three methods (except against FGSM) while achieving higher accuracy on benign images. Since we design an identical optimized table for all three channel instead of different table for each channel in traditional JPEG, we use our quantization table in next subsection.

\begin{table*}[htbp]
	\renewcommand\arraystretch{1}
	\centering
	\caption{Performance of image classifiers trained with adversarial training and proposed method.}
	\begin{tabular}{ccccccc}
		Method& $\epsilon$ &Epochs&Standard acc.&PGD + 1 restarts&PGD + 10 restarts&Total time(hrs) \\ 
		\hline
		Fast \cite{at_fast} &2&15&60.90\%&43.46\%&43.43\%&12.14 \\
		Free \cite{at_free} &2&15&64.37\%&43.31\%&43.28\%&52.20 \\
		Proposed&2&14&65.10\%&40.06\%&40.01\%&11.51\\
		Proposed&2&28&63.10\%&42.23\%&42.17\%&22.11\\
		\hline
		Fast \cite{at_fast} &4&15&55.45\%&30.28\%&30.18\%&12.14 \\
		Free \cite{at_free} &4&15&60.42\%&31.22\%&31.08\%&52.20 \\
		Proposed&4&14&64.02\%&28.43\%&28.40\%&11.51\\
		Proposed&4&28&63.01\%&30.89\%&30.85\%&22.11\\
		\hline
		
	\end{tabular}
	\vspace{-3mm}
	\label{tab: 01}
\end{table*}

\subsection{Noisy Training}
\label{ae_training}

At present, the vanilla adversarial training is that, in each iteration, the network is trained with those adversarial examples generated online \cite{denoise,mmat}. In this way, the trained network can defend against the attack of the sample to a certain extent. However, there are two disadvantages in traditional adversarial training: firstly, there is no theoretical guarantee for this kind of defense, that is, we do not know whether the attacker can design more intelligent attack methods to bypass this defense. Secondly, this kind of adversarial training can be extremely time-consuming since the calculation of adversarial examples is quite complicated. 


More recently,  adversarial training \cite{at_fast} proposed to accelerate training from three aspects, (1) starting from a non-zero initial perturbation, regardless of the actual initialization. (2) one of the reasons why FGSM and R+FGSM \cite{R+FGSM}  may have failed previously is due to the restricted nature of the generated examples (each dimension is  perturbed only by either 0 or $\pm$$\epsilon$). 
(3) defenders do not need strong adversaries during training.
Thus, fast adversarial training (FAT) is designed to modify the simple FGSM (instead of more complicated PGD \cite{madry2018towards}) adversarial training by replacing its zero initial with random initial. However, since we do not need strong adversaries, why not to use a simpler Gaussian noise as adversaries? Also, \cite{gaussian_adv, gaussian_adv2} have suggested that Gaussian data augmentation could supplement or replace adversarial training and  proposed using a network’s robustness to Gaussian noise as a proxy for its robustness to adversarial perturbations. Inspired by this, we proposed a noisy training based method.  With just an adding operation (without computation of gradients), we consider our training computation complexity is a little less than that of \cite{at_fast}. Specifically, noisy based adversarial training combined with our method uses compression  as the regularizer for original cost function, which is as follows
\begin{equation}
J'(\theta,x,y) = \xi J(\theta,x,y) + (1-\xi) J(\theta,x_{q},y)
\end{equation}
where $\xi \in [0,1]$. The $x_{q}$ represents compressed image at quality $q$ of original image $x$ adding Gaussian noise $\rho_{g}$. In our experiments, we use $\xi=0.9$ to achieve diversity of networks' parameters \cite{diversity}.
Note that the proposed noisy training is different from randomized smoothing proposed in \cite{lecuyer2019certified,cohen2019certified,zheng2019unified}, where Gaussian noise is injected into original images rather than the compressed images as done in this paper. As will be verified by our experiments later, combining noisy training with compression based transformation can provide both
$L_{2}$ and $L_{\infty}$ robustness.

\begin{figure}[t]
	\includegraphics[scale=0.65]{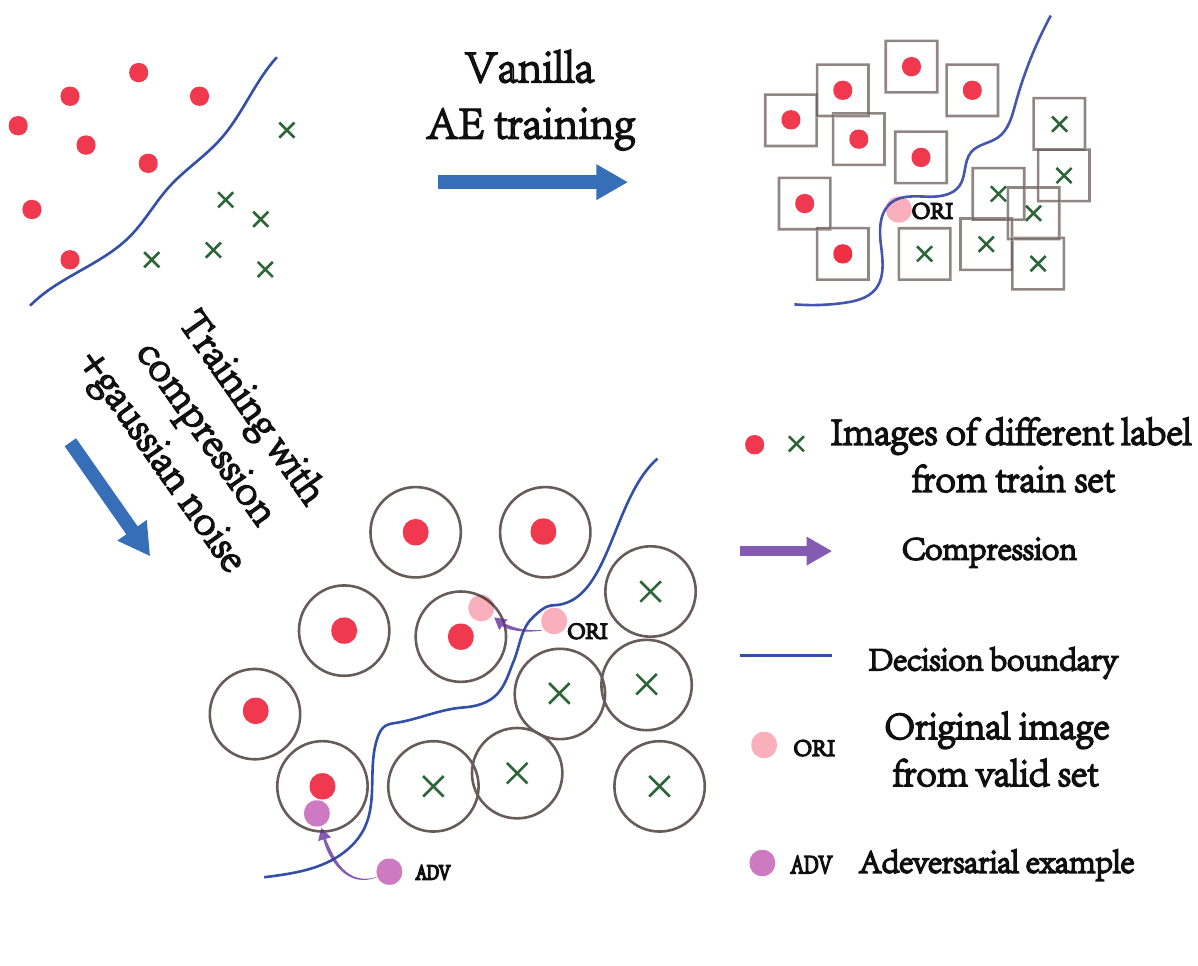}
	\caption{Comparison between vanilla adversarial example (AE) training and the proposed training algorithm. }
	\label{fig:07}
	\vspace{-5mm}
\end{figure}

\par
We adopt the method in \cite{jpg2,shield} to fine-tune the network using the pre-trained weights for faster convergence. We use stochastic gradient descent (SGD) with a learning rate of 0.005, with a decay of 94\% over 14 / 28 epochs. Training images are compressed at qualities from 90 to 30 (with a step size of 10). Hence we get a total of 8 network models sharing the same architecture but having different weights ($M, M_{90}, M_{80}, M_{70}, M_{60}, M_{50}, M_{40}, M_{30}$), where $M$ represents the original network model and $M_{x}$ the network model fine-tuned with images of compression quality of $x$. While fine-tuning, the initial weights of $M_{x}$ is from the intermediately proceeding one, i.e.,   $M_{x+10}$, and the initial weights of $M_{90}$ is from the initial pre-trained model $M$. For an ablation study, our proposed noisy training is  compared with Fast \cite{at_fast} and Free \cite{at_free} training in terms of computation efficiency and robust accuracy, and the results are listed in Table \ref{tab: 01}. We can see the proposed method maintains higher accuracy on benign image (Standard acc.) than Fast training and has similar robustness to PGD attack. Compared with Free training, we achieve the same level of performance while consuming much less time. 

Further, we test the performance of each model in our model set. Figure \ref{fig:06} shows the results of classification accuracy on benign images and adversarial examples. In Figure \ref{fig:06} (a), the results on benign images show fine-tuning with certain compression quality images can indeed improve network's accuracy. However, this improvement is limited to certain compression quality that was used for training, e.g., $M_{90}$ achieves best performance on images at quality 90 (Q90).  Despite those network models fine-tuned in later stage ($M_{50}, M_{40}, M_{30}$) have been learned from high quality images, those models do not perform well on high quality images. For example, the $M_{50}$ is indirectly fine-tuned from the $M$, but $M_{50}$ performs bad on original images. In Figure \ref{fig:06} (b), as the compression ratio increases, fine-tuned models with compressed images all can achieve better performance on defending adversarial examples than the original model $M$. But defense efficiency becomes better when compression quality is lower. So, using one single network to model seems to be difficult to maintain good performance on both high and low quality images. To save inference time, finally we choose $M, M_{90}, M_{70}, M_{50}, M_{30}$ to consist our network model set which is used as an ensemble for defense.

Finally, we use  a example, i.e., Figure \ref{fig:07}, to geometrically illustrate the key difference between the   vanilla adversarial training and the proposed training method. As studied in \cite{at_fast}, vanilla adversarial training can lead to ``catastrophic overfitting'', which improves robustness, however, at the cost of accuracy on original test images. This issue is shown in the top-right of Figure \ref{fig:07}. In contrast, our proposed training method can improve both accuracy on benign images and defense efficiency  on adversarial examples  as verified in Figure \ref{fig:06}. This means that, our proposed method can correctly recognize more image examples. That is, some original images and adversarial examples could be pushed back to correct decision region, as shown in the bottom of Figure \ref{fig:07}. It should be noted here, although combined with compression and Gaussian nosie, our proposed method may also have the problem of overfitting.


\section{Performance Evaluation}

\subsection{Experiment Setup}
Our experiments are conducted on the Tensorflow DNN computing framework \cite{tensorflow}. We choose the large-scale ImageNet \cite{imagenet} dataset as our benchmark to better illustrate the proposed method. Five types of adversarial example attacks including FGSM \cite {fgsm}, I-FGSM \cite{bim}, Deepfool \cite{deepfool}, C$\&$W($L_{2}$) \cite{cw} and BPDA \cite{bpda}, are simulated using popular cleverhans package on validation set of ImageNet (50k images) in our experiment for evaluating the efficiency of each defense. Feature Distillation (FD) \cite{fd} and Total variance minimization (TVM) \cite{tvm&quilting} are selected as our defense benchmarks to compare with our proposed method. To the best of our knowledge, these two methods yield the state-of-the-art performance on countering the adversarial attack among model-agnostic methods. We adopt the ResNet-v2-50 \cite{resnet1, resnet2} for Gray-box and Black-box setting, Inception-v4 \cite{incep} for White-box setting.

\begin{table*}[tbp]
	\renewcommand\arraystretch{1.1}
	
	\centering
	\caption{Summary of model accuracy (in \%) for all defenses in Gray-box and Black-box scenarios.}
	\label{tab: 02}
	\small
	\begin{center}
		\begin{tabular}{c|c|c|c|c|c|c|c|c|c|c|c}
			\toprule[2pt]
			\multirow{2}{*}{\textbf{Gray-box}} &\multirow{2}{*}{No attack} & \multicolumn{3}{|c|}{FGSM \cite{fgsm}} & \multicolumn{3}{c|}{I-FGSM \cite{bim}}&\multicolumn{2}{c|}{\multirow{2}{*}{Deepfool \cite{deepfool}}}&\multicolumn{2}{c}{\multirow{2}{*}{C\&W($L_{2}$) \cite{cw}}} \\
			&&\multicolumn{1}{|c|}{$\epsilon=2$}&$\epsilon=4$&\multicolumn{1}{c|}{$\epsilon=8$}&$\epsilon=2$&$\epsilon=3$&\multicolumn{1}{c|}{$\epsilon=4$}&\multicolumn{2}{c|}{}& \multicolumn{2}{}{}\\
			\hline
			No Defense  &65.4&17.5&14.4&13.2&15.1&12.4&9.5&\multicolumn{2}{c|}{9.7}&\multicolumn{2}{c}{9.3}\\
			\hline
			JPEG (75) \cite{jpeg}    &60.5&41.1&29.4&21.1&57.1&51.6&50.0&\multicolumn{2}{c|}{60.0}&\multicolumn{2}{c}{60.3} \\
			\hline
			FD \cite{fd}         &59.5&44.1&30.5&21.4&56.5&54.4&53.3&\multicolumn{2}{c|}{58.8}&\multicolumn{2}{c}{58.7} \\
			\hline
			TVM \cite{tvm&quilting}        &58.2&44.7&33.7&22.7&55.1&51.9&48.8&\multicolumn{2}{c|}{57.6}&\multicolumn{2}{c}{58.8}\\
			\hline
			Proposed&\textbf{66.5}&\textbf{56.4}&\textbf{43.1}&\textbf{28.2}&\textbf{68.0}&\textbf{65.2}&\textbf{64.8}&\multicolumn{2}{c|}{\textbf{67.1}}&\multicolumn{2}{c}{\textbf{65.4}}\\
			
			\toprule[2pt]
			\multirow{2}{*}{\textbf{Black-box}} &\multirow{2}{*}{No attack} & \multicolumn{3}{|c|}{FGSM \cite{fgsm}} & \multicolumn{3}{c|}{I-FGSM \cite{bim}}&\multicolumn{2}{c|}{TI-FGSM \cite{dong2019evading}}&\multicolumn{2}{c}{DI-FGSM \cite{dim}}\\
			&&\multicolumn{1}{|c|}{$\epsilon=2$}&$\epsilon=4$&\multicolumn{1}{c|}{$\epsilon=8$}&$\epsilon=2$&$\epsilon=3$&\multicolumn{1}{c|}{$\epsilon=4$}&\multicolumn{1}{c|}{$\epsilon=2$}&\multicolumn{1}{c|}{$\epsilon=4$} &\multicolumn{1}{c|}{$\epsilon=2$}&\multicolumn{1}{c}{$\epsilon=4$}\\
			\hline
			No Defense&65.4&45.0&34.6&25.3&53.5&55.8&54.7&\textbf{56.1}&\textbf{48.4}&45.6&38.4\\
			\hline
			JPEG (75) \cite{jpeg}  &60.5&50.9&40.1&28.7&55.8&56.8&56.2&52.9&47.0&51.4&42.3 \\
			\hline
			FD \cite{fd}       &59.5&50.9&41.2&28.5&56.5&56.7&55.6&51.3&46.9&50.4&44.1\\
			\hline
			TVM \cite{tvm&quilting}      &58.2&59.3&42.9&32.6&52.3&55.7&55.7&51.5&47.0&\textbf{52.1}&45.1\\
			\hline
			Proposed&\textbf{66.5}&\textbf{62.1}&\textbf{50.7}&\textbf{34.9}&\textbf{62.0}&\textbf{63.1}&\textbf{66.7}&55.1&48.0&52.0&\textbf{45.3}\\
			\toprule[2pt]		
			
		\end{tabular}
	\end{center}
	\vspace{-7mm}
\end{table*}

%

\subsection{Gray-box Scenario}
In this scenario, the adversary has the access to the model's architecture and parameters but is unaware of defense strategy. We use the parameters of the model which is trained with benign images to generate adversarial examples. As Table \ref{tab: 02} shows, when facing with benign images, the proposed method can maintain accuracy at the same level as original model. Traditional JPEG, FD \cite{fd} and TVM  \cite{tvm&quilting} all introduce 5\% drop on benign image (No attack). When attacking, Deepfool and C\&W  are both very effective,  because they can achieve beyond $90\%$ success rate while maintaining a much lower $L_{2}$ dissimilarity with original image. However, these two low intensity but effective attacks can be defended very largely by all four transformation defense. Moreover, our proposed methods can achieve the defense efficiency as high as benign images' accuracy.  When combating with FGSM \& I-FGSM, which generate adversarial examples with stronger perturbation intensity, our proposed methods achieve around 10\% higher performance than other three methods.

\subsection{Black-box Scenario}
In this scenario, the adversary is unaware of neither the architecture nor the parameters about the network. In this setup, we intend to evaluate the transferability of attacked images which are generated using ResNet-v2-101 and tested on ResNet-v2-50. Because the attacks with extremly small perturbation magnitude like Deepfool and C\&W usually have limited transferability, we adopt recently proposed black-box attack methods \cite{dim} and \cite{dong2019evading} into evaluation. Compared to employing the same attack parameters  in Gray-box setting, adversarial examples have certain degree of transferability. As we can see in Table \ref{tab: 02}, though these adversarial examples have weak effect on networks, traditional JPEG, FD  and TVM  can hardly defend these transferred adversarial examples, especially for stronger perturbation intensity. Accuracy even declines when facing TI-FGSM \cite{dong2019evading} attack. As can be seen, TI attacks actually have worse transferability performance than FGSM. This is probably because the TI attack may be counterproductive if the discriminative regions of normally trained model and defense models are not that different in the case of low-strength attack. 
Further, the attack algorithm should generate adversarial examples with smaller dissimilarity while adversarial perturbations with $\epsilon=8,16$ as set in \cite{dong2019evading} have poor visual quality, which can be easily identified by human eyes. So we did not consider them in test. When facing larger intensity FGSM attacks, our proposed method can still achieve 8\% higher accuracy than other defenses.

\subsection{White-box Scenario}
In this scenario, we evaluate our methods against white-box BPDA attack, which generates adversarial examples using defense methods iteratively. We implement the evaluation experiment on the released code at GitHub \cite{cw}, using Inception-v4 \cite{incep} as test model, and 1000 iterations and 0.1 learning rate as attack parameters. The accuracy of various methods on adversarial examples ($Acc_{ae}$ ) and benign images ($Acc_{raw}$) are reported in Table \ref{tab: 03}. As can be observed, BPDA attack almost achieves 100\% success rate. Due to our multiple-model setting, we can defend BPDA attack to a large extent. This is because BPDA works by finding a differentiable approximation ($f(\cdot)$) for a non-differentiable preprocessing transformation (\emph{e.g.}, quantization) or network layer, $g(\cdot)$. Apparently, our proposed ensemble model set has various non-differentiable $g(\cdot)s$, and  there would be extremely difficult for BPDA to find \emph{one} $f(\cdot)$ to simultaneously approximate our \emph{five} different $g(\cdot)s$, if not impossible. So  our method is an easy and effective way to defend BPDA attack, and provides a new angle to redsign input-based defense to balance the accuracy of benign image and defense efficiency with defensive frequency domain quantization.

\begin{table}[htbp]
	\renewcommand\arraystretch{1}
	\centering
	\caption{Accuracy results (in \%) in white-box scenario. }
	\begin{tabular}{cccccc}
		\toprule[2pt]  
		&None &JPEG (75) &FD &TVM &Proposed \\ 
		\midrule[2pt]  
		$Acc_{raw}$ &78&74&73&74&78 \\
		$Acc_{ae}$  &0&1&2&0&60 \\
		\bottomrule[2pt]  
	\end{tabular}
	\label{tab: 03}
\end{table}

\section{Conclusion}
In this paper, we combine the compression based input transformation method with re-training the compressed images. Firstly, we propose an optimized JPEG compression process to remove perturbations as much as possible and preserve the original features at the same time, by discarding the color space transformation and introducing an iterative algorithm to optimize the quantization table. Secondly we use images compressed with certain level as a strategy of data augmentation and  generate a set of different network models. The final prediction will be the average results voted by each model in the set. Experimental results show that, our proposed method can improve defense efficiency while maintaining the  classifying accuracy for benign images. 

\clearpage

\small
\bibliographystyle{ieee_fullname}
\bibliography{egbib}

\end{document}